%% file: RGB-T_Segmentation.tex
\documentclass[preprint,12pt,3p]{elsarticle}

\usepackage{bbding}
\usepackage{array}
\usepackage{verbatim}

\usepackage{amsmath,amssymb,amsfonts}
\usepackage{times}
\usepackage{epsfig}
\usepackage{graphicx}
\usepackage{subfigure}
\usepackage{multirow}
\usepackage{color}
\usepackage{natbib}
\usepackage{caption}
\usepackage{bm}
\usepackage{bbm}
\usepackage{mathptmx}  
\usepackage{mathrsfs}
\usepackage{amsthm}
\usepackage{amsfonts}
\usepackage{url} 
\usepackage{textcomp}
\usepackage[pagebackref=true,breaklinks=true,letterpaper=true,colorlinks,bookmarks=false,citecolor=blue]{hyperref}
\usepackage{booktabs}
\usepackage{tabu}
\usepackage{color}
\usepackage[abs]{overpic}
\usepackage[linesnumbered,ruled,vlined]{algorithm2e}
\usepackage{algpseudocode}
\usepackage[table]{xcolor}

\begin{document}

\begin{frontmatter}

\title{Channel and Spatial Relation-Propagation Network for RGB-Thermal Semantic Segmentation}

\author[label1,label3]{Zikun Zhou\corref{cor2}}
\author[label3]{Shukun Wu\corref{cor2}}
\author[label3]{Guoqing Zhu}
\author[label3,label1]{Hongpeng Wang}
\author[label1,label3]{Zhenyu He\corref{cor1}}\ead{zhenyuhe@hit.edu.cn}
\address[label1]{Peng Cheng Laboratory, Shenzhen 518055, China.}
\address[label3]{School of Computer Science and Technology, Harbin Institute of Technology, Shenzhen 518055, China}
\cortext[cor2]{Zikun Zhou and Shukun Wu contribute equally.}
\cortext[cor1]{Zhenyu He is the corresponding author.}

\begin{abstract}
\input{abstract}
\end{abstract}


\end{frontmatter}


\section{Introduction}
\input{introduction}

\section{Related Work}
\input{related_work}

\section{Method}
\input{method}

\section{Experiments}
\input{experiments}

\section{Conclusion}
\input{conclusion}

\section*{Acknowledgements}
\input{acknowledgements}

\bibliographystyle{elsarticle-num} 
\bibliography{rgbtsegbib}

\end{document}

%% file: abstract.tex
RGB-Thermal (RGB-T) semantic segmentation has shown great potential in handling low-light conditions where RGB-based segmentation is hindered by poor RGB imaging quality. The key to RGB-T semantic segmentation is to effectively leverage the complementarity nature of RGB and thermal images. Most existing algorithms fuse RGB and thermal information in feature space via concatenation, element-wise summation, or attention operations in either unidirectional enhancement or bidirectional aggregation manners. However, they usually overlook the modality gap between RGB and thermal images during feature fusion, resulting in modality-specific information from one modality contaminating the other. In this paper, we propose a Channel and Spatial Relation-Propagation Network (CSRPNet) for RGB-T semantic segmentation, which propagates only modality-shared information across different modalities and alleviates the modality-specific information contamination issue. Our CSRPNet first performs relation-propagation in channel and spatial dimensions to capture the modality-shared features from the RGB and thermal features. CSRPNet then aggregates the modality-shared features captured from one modality with the input feature from the other modality to enhance the input feature without the contamination issue. While being fused together, the enhanced RGB and thermal features will be also fed into the subsequent RGB or thermal feature extraction layers for interactive feature fusion, respectively. We also introduce a dual-path cascaded feature refinement module that aggregates multi-layer features to produce two refined features for semantic and boundary prediction. Extensive experimental results demonstrate that CSRPNet performs favorably against state-of-the-art algorithms.

%% file: introduction.tex
Semantic segmentation, which aims to predict the category for every pixel in an image, is an important task in computer vision with a diverse range of applications, such as autonomous driving~\cite{DepthAdaNet,GMNet,BASeg}, medical diagnosis~\cite{Unet,PyDiNet,MultiResUNet}, and robot sensing~\cite{RAttSeg}. Recently, many deep segmentation algorithms~\cite{DANet,FaPN,UCTransNet} have made significant progress in RGB-based semantic segmentation. However, the imaging quality of visible light cameras degenerates in low-light conditions, restricting the application of the RGB segmentation methods in such extreme conditions.

To enhance segmentation robustness in extreme conditions, various studies~\cite{GMNet,RDFNet,DepthAwareCNN,ABMDRNet} utilize other modality data to complement RGB images. Commonly used additional modalities include depth~\cite{RDFNet,DepthAwareCNN,SAG} and thermal~\cite{GMNet,ABMDRNet,RTFNet} images. Depth images can provide 3D geometric information of the scene, while thermal images measure the thermal radiation of any object whose temperature is over absolute zero. However, depth cameras utilizing structured light or time-of-flight technology remain vulnerable to strong outdoor lighting conditions, thus limiting their application in extreme outdoor situations. By comparison, thermal cameras are less susceptible to lighting conditions and work well under all weather. Hence, we focus on RGB-Thermal (RGB-T) semantic segmentation in this work. The common characteristics of RGB and thermal images are that they both reflect the shape of objects in the scene. The specific characteristics lie in that RGB images depict color distribution while thermal images capture temperature distribution.

\begin{figure*}[t]
\centering
\includegraphics[width=1\linewidth]{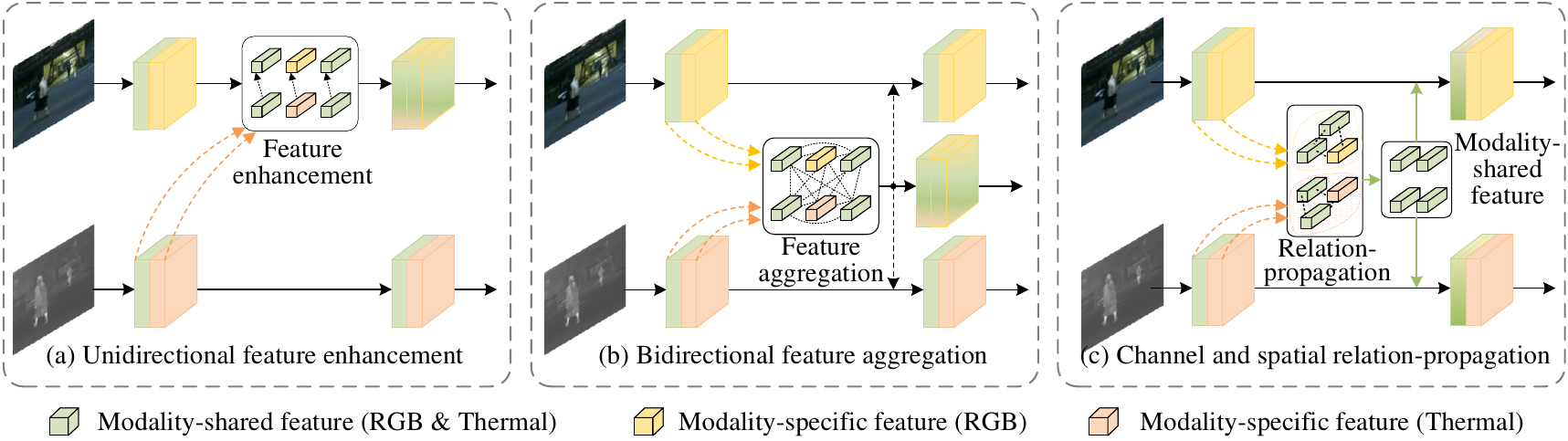}
\vspace{-2mm}
\caption{\textbf{Illustration of different feature fusion methods for RGB-T semantic segmentation.} (a) Unidirectional feature enhancement~\cite{RTFNet,FEANet}, which uses thermal features as auxiliary information to enhance RGB features via element-wise summation or concatenation. (b) Bidirectional feature aggregation~\cite{ABMDRNet,EGFNet}, which treats RGB and thermal images equally and aggregates the multi-modality features via element-wise summation or concatenation with sophisticated attention operations. (c) Our proposed channel and spatial relation-propagation, which focuses on capturing and interactively fusing the modality-shared features.}
\label{Fig:introduction}
\vspace{-4mm}
\end{figure*}

The core challenge in RGB-T segmentation lies in how to effectively fuse multi-modality data to make them complementary to each other. A common pipeline for multi-modality fusion is to use a two-stream network to extract deep features of different modalities and then fuse the deep features~\cite{GMNet,ABMDRNet,RTFNet,Fusenet,MFFENet,ACNet}. Among these methods, several algorithms~\cite{RTFNet,FEANet,Fusenet} employ a unidirectional feature enhancement strategy for multi-modality feature fusion. As shown in Figure~\ref{Fig:introduction}~(a), these methods use the thermal feature as auxiliary information to enhance the RGB feature via element-wise summation or concatenation. Following this, they perform segmentation based on the enhanced feature. However, using the thermal image as an auxiliary modality is less effective for night scenes, as the thermal image plays a more crucial role in revealing night scenes compared to the RGB image.

Besides, numerous algorithms~\cite{GMNet,RDFNet,ABMDRNet,EGFNet,ACNet,CRF} utilize the bidirectional feature aggregation strategy for multi-modality fusion. As shown in Figure~\ref{Fig:introduction}~(b), they treat the RGB and thermal images equally in multi-modality feature fusion, usually involving sophisticated attention mechanisms to improve the feature aggregation. For example, several approaches~\cite{GMNet,ABMDRNet,EGFNet} first aggregate the features from different modalities via element-wise summation or concatenation. Then they use the attention operation to model long-range dependencies to enhance the aggregated features. Generally, feature aggregation is independent of feature extraction~\cite{GMNet,RDFNet,ACNet,ABMDRNet,EGFNet}. In particular, several algorithms~\cite{SAG,NANet,CMX} attempt to aggregate the multi-modality features interactively. Namely, they feed the aggregated features into the feature extraction block of the next layer, as depicted by dashed arrows in Figure~\ref{Fig:introduction}~(b). Although the above algorithms achieve advanced performance, they overlook the discrepancies between the modality-specific characteristics of RGB and thermal images (also named modality gap) during fusion. Especially for interactive aggregation methods, aggregating the modality-specific information of one modality into the other modality will contaminate the specific characteristics of the latter.

In this paper, we propose a Channel and Spatial Relation-Propagation Network (CSRPNet) for RGB-T semantic segmentation. The core idea is to first capture the features containing modality-shared information (also called modality-shared features) and then utilize them to perform interactive multi-modality fusion, as shown in Figure~\ref{Fig:introduction}~(c). By aggregating only the modality-shared features with the original features, we can alleviate the issue of modality-specific information from one modality contaminating the other during interactive fusion. 

CSRPNet captures the modality-shared features from the RGB and thermal features via relation-propagation in channel and space. The rationale behind this design is: (1) the inter-channel and inter-pixel relation matrices can be used to represent the intra-modality semantic dependencies at the feature channel and pixel levels, respectively; (2) and thus the feature channels or pixels sharing similar relations across different modalities are more likely to model modality-shared information. In this way, we can obtain the modality-shared feature channels and pixels by capturing those sharing similar relations. Such a process is named relation-propagation. Herein, we investigate two relation functions, including dot product and Gaussian function, to model the relation between two channels or two pixels for calculating the relation matrices. By integrating captured modality-shared channels and pixels from one modality with the other modality, we can adaptively and interactively fuse multi-modality features while avoiding feature contamination resulting from the modality gap. Based on the above idea, we design a channel and spatial relation-propagation module and plug it into different layers of the two-steam backbone for interactive fusion.

To fully exploit the multi-layer fusion features, we design a Dual-path Cascaded Feature Refinement (DCFR) module. The DCFR module aggregates the multi-layer fusion features layer-by-layer via two paths, generating two refined features for accurate semantic prediction and boundary prediction, respectively. To conclude, we make the following contributions:

\begin{itemize}
\item We propose a channel and spatial relation-propagation network for RGB-T segmentation. It can fuse the multi-modality features interactively while preserving modality-specific information for each modality by capturing and integrating the modality-shared features.

\item We propose a channel and spatial relation-propagation module with two types of relation functions, which enables the effective capture of modality-shared features. We also design a dual-path cascaded feature refinement module to fully utilize the multi-layer fusion features.

\item We achieve favorable performance against state-of-the-art methods on the MFNet and PST900 datasets, demonstrating the effectiveness of the proposed algorithm.
\end{itemize}

%% file: related_work.tex
\subsection{RGB Semantic Segmentation}
Significant progress has been achieved in semantic segmentation with the development of deep learning-based algorithms~\cite{Unet,DANet,FCN,DUC,arora2023fractional}.
The pioneering method FCN~\cite{FCN} proposes a simple yet effective framework for dense segmentation, which is capable of accepting images of any size as input and substantially improves the segmentation efficiency. Despite the great progress, the prediction mask of the FCN method tends to lose fine details. Numerous methods~\cite{Unet,LDN,DUC} have been proposed to address this issue. For example, U-Net~\cite{Unet} fuses the feature map in the encoder with the feature map in the corresponding layer of the decoder to obtain richer context information to improve the segmentation performance. DUC~\cite{DUC} proposes the dense upsampling convolution operation to replace the original bilinear upsampling and deconvolution operations. To reduce computational and memory loads, SegNet~\cite{Segnet} proposes the max pooling indices to perform the non-linear upsampling of the feature maps in the decoder. Building on the success of DenseNet~\cite{DenseNet} on image classification, FC-DenseNet~\cite{FC-DensetNet} adapts and extends the DenseNet architecture into a fully convolutional network for semantic segmentation.

Recently, several methods~\cite{DANet,ccnet} exploit the attention mechanism to capture more contextual information for semantic segmentation. CCNet~\cite{ccnet} proposes a criss-cross attention module to model contextual information from surrounding pixels on the cross path. DANet~\cite{DANet} introduces a position attention module to selectively aggregate the long-range contextual information according to the spatial attention map. It also proposes a channel attention module to model inter-dependencies between different semantic responses based on the channel attention map. Despite the excellent performance, these RGB-based semantic segmentation methods remain highly susceptible to lighting conditions.
 
\subsection{Multi-modality Semantic Segmentation}
In recent years, multi-modality data have been utilized for semantic segmentation to tackle the issue that the RGB segmentation methods are vulnerable to lighting conditions. And numerous RGB-Depth (RGB-D)~\cite{RDFNet, DepthAwareCNN, SAG, Fusenet, ACNet, NANet, CMX, SCN, pavel2017object} and RGB-T~\cite{GMNet, ABMDRNet, RTFNet, FEANet, EGFNet, MFFENet, CRF, MFNet, FuseSeg} segmentation algorithms have been proposed. Multi-modality feature fusion is the core challenge for multi-modality semantic segmentation. Existing multi-modality feature fusion methods can be coarsely divided into two categories: unidirectional feature enhancement and bidirectional feature aggregation.

The methods~\cite{RTFNet, FEANet, Fusenet, MFFENet, FuseSeg} employing the unidirectional feature enhancement strategy leverage the depth or thermal image as auxiliary information to enhance RGB information. Specifically, they usually feed the deep features extracted by the depth or thermal backbone into the RGB backbone to enhance the RGB features for robust semantic segmentation. RTFNet~\cite{RTFNet} and FEANet~\cite{FEANet} use two independent ResNet~\cite{ResNet} models as backbones to extract the deep features of the RGB and thermal images. To enhance the RGB features with the thermal features, they integrate the corresponding thermal features into the RGB backbone stream via element-wise feature summation. The unidirectional feature enhancement strategy allows RTFNet and FEANet to perform semantic segmentation directly based on the RGB backbone stream features. However, such a strategy is less effective in processing night scenes, due to the thermal image playing a more important role than the RGB image.

Algorithms~\cite{GMNet, RDFNet, ABMDRNet, SAG, EGFNet, ACNet, CRF, NANet, CMX} exploiting the bidirectional feature aggregation strategy treat data from different modalities equally in aggregation. Several methods~\cite{ABMDRNet, ACNet} leverage attention-based fusion modules to aggregate multi-modality features. For example, ACNet~\cite{ACNet} utilizes a channel attention-based module to extract the weighted RGB and depth features and then aggregates them via element-wise summation. ABMDRNet~\cite{ABMDRNet} proposes a multi-scale spatial and channel context module to capture long-range dependencies along the spatial and channel dimensions to improve RGB-T segmentation accuracy. Some methods~\cite{GMNet, RDFNet, EGFNet} use customized fusion modules to obtain discriminative cross-modality features. Specifically, RDFNet~\cite{RDFNet} first uses residual convolution units to refine the RGB and thermal image features separately and then merge the refined features by element-wise summation. GMNet~\cite{GMNet} divides the multi-stage features into ``senior", ``intermediate", and ``junior" features, and further introduces the shallow and deep feature fusion modules at different stages to merge RGB and thermal features.

Unlike the above methods, the other methods~\cite{SAG, NANet, CMX} utilizing the bidirectional feature aggregation strategy aggregate the features in an interactive manner. They usually feed the aggregated features back to the next feature extraction layer of the backbone, aiming to maximize cross-modality synergies. Specifically, SAG~\cite{SAG} proposes an SA-Gate to perform informative feature propagation between different modalities via channel-wise attention. NANet~\cite{NANet} employs the non-local~\cite{non-local} operation to interactively exchange information between the RGB and depth modalities along the spatial and channel dimensions. CMX~\cite{CMX} proposes a transformer-based multi-modality fusion architecture for RGB-X semantic segmentation. It introduces a cross-modal feature rectification module to conduct channel-wise and spatial-wise rectification, and a feature fusion module to merge the multi-modality features. However, this directly interactive fusion manner without considering the modality gap inevitably results in contaminating the specific information of each modality. Differently, our proposed model first mines the modality-shared features and then interactively fuse the captured modality-shared features, alleviating the modality-specific information contamination issue.

%% file: method.tex
\begin{figure}[ht] 
\centering 
\includegraphics[width=1.0\textwidth]{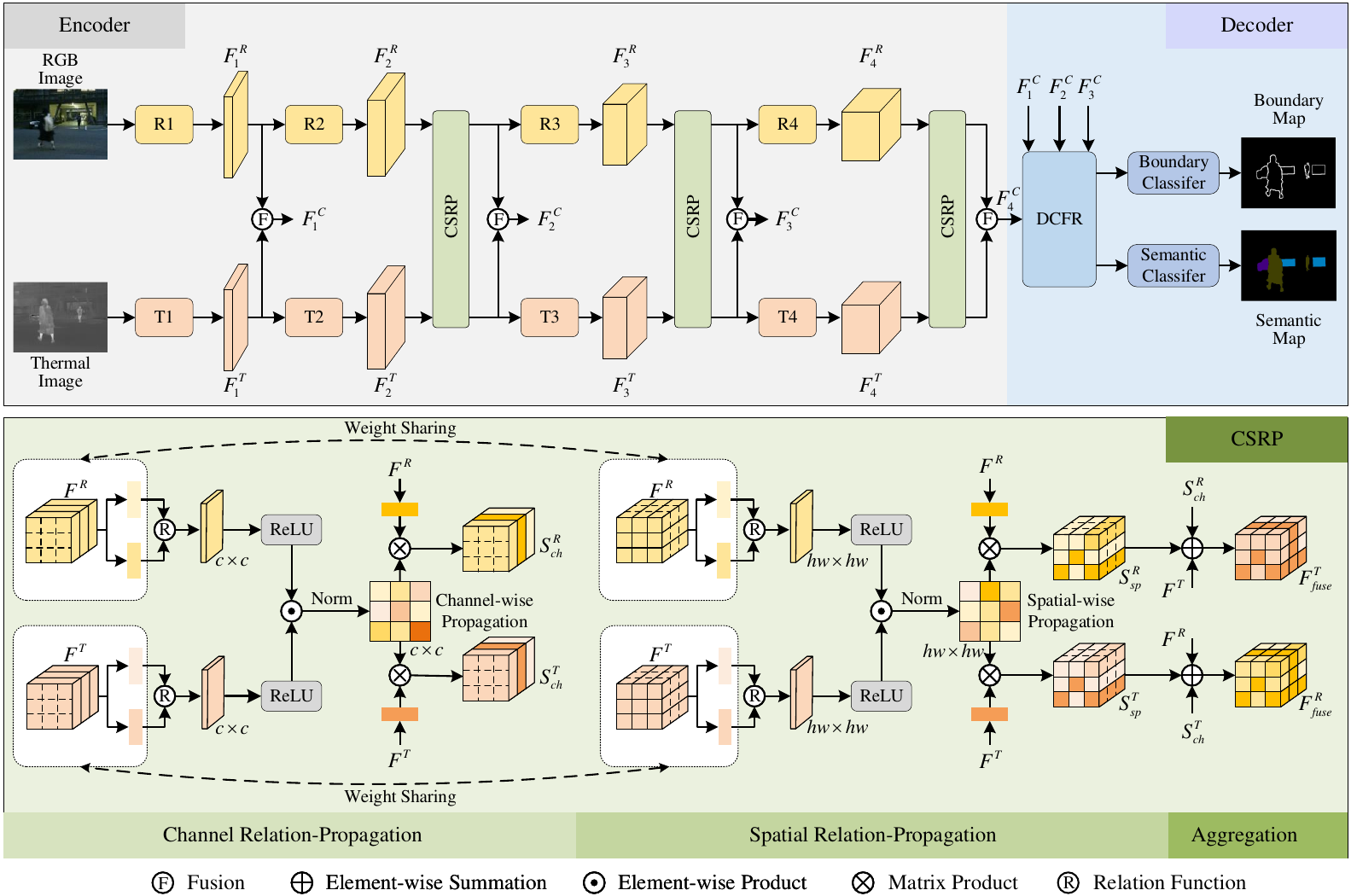} 
\caption{\textbf{Architecture of proposed CSRPNet for RGB-T semantic segmentation.} CSRPNet takes as input the RGB and thermal images, and outputs the predicted boundary map and semantic map. The proposed CSRP module performs relation-propagation in channel and spatial to extract the modality-shared feature and then fuse the multi-modality feature interactively. The DCFR module takes as input the multi-layer features and outputs two refined feature maps for boundary classification and semantic classification.}
\label{Fig:framework}
\end{figure}

The crux of RGB-T semantic segmentation is to overcome the modality gap between the RGB and thermal images and to effectively fuse their features.
To surmount this crux, we propose a Channel and Spatial Relation-Propagation Network (CSRPNet), which first captures the modality-shared features via relation-propagation in the channel and spatial dimensions and then aggregates the captured modality-shared features.
Figure~\ref{Fig:framework} illustrates the overall architecture of CSRPNet, which comprises an encoder and a decoder. The encoder consists of a two-stream feature extraction network and the Channel and Spatial Relation-Propagation (CSRP) module. The encoder takes as input the RGB and thermal images and extracts their deep features while fusing the multi-modality features in an interactive manner. The decoder is comprised of a Dual-path Cascaded Feature Refinement (DCFR) module, a boundary classifier, and a semantic classifier; it takes as input the multi-layer fused features and outputs the boundary and segmentation maps.
Here, learning to predict the boundary map can enhance the ability of the model to perceive boundary details, which in turn improves the semantic segmentation precision. In the following, we present each component of CSRPNet in detail.

\subsection{Two-stream feature extractor and interactive fusion}
We use a two-stream backbone network to extract the feature of the RGB and thermal images. Similar to~\cite{RTFNet,EGFNet,FEANet}, we use ResNet-50~\cite{ResNet} pre-trained on ImageNet~\cite{imagenet} as the backbone. Note that the RGB and thermal backbones share a similar structure, with the only difference being that the input channel of the thermal backbone is modified to 1 since the thermal image is single-channel. Given the RGB image $I^{R}$ and thermal image $I^{T}$, the two-stream backbone network extract the multi-layer RGB features $\{F^{R}_l|_{l=1, 2, 3, 4}\}$ and the multi-layer thermal features $\{F^{T}_l|_{l=1, 2, 3, 4}\}$. 
To effectively leverage the complementarity between RGB and thermal images, we adopt an interactive feature fusion manner. Specifically, we first enhance the feature from each modality with the feature from the other one using the CSRP module without contaminating modality-specific information. Then we feed the enhanced RGB or thermal features back to the subsequent feature extraction layer while fusing them together. This interactive fusion manner enables the deeper feature extraction layer to accept and process information from both RGB and thermal images, promoting tight coupling between multi-modality feature extraction and fusion. In the following, we present the channel and spatial relation-propagation module for multi-modality feature fusion.

\subsection{Channel and spatial relation-propagation}
The modality gap between the RGB and thermal images presents a problem in directly fusing their features interactively: directly aggregating the RGB and thermal features and feeding them back into the two-stream backbone network could lead to the contamination of modality-specific information in each modality. To handle the problem, we develop a Channel and Spatial Relation-Propagation (CSRP) module to capture and aggregate the modality-shared features from different modalities. The modality-shared features enable interactive feature fusion without contaminating the modality-specific feature. The CSRP module is composed of a Channel Relation-Propagation (CRP) block, a Spatial Relation-Propagation (SRP) block, and an aggregation block.

The CRP and SRP blocks aim to capture the modality-shared features for interactive multi-modality feature fusion by considering the channel-wise and pixel-wise relations, respectively. For this purpose, we first compute the relation matrices to represent the semantic and spatial dependencies between different channel pairs and pixel pairs, respectively. Then we capture feature channels and pixels sharing common relations across different modalities via relation-propagation. Specifically, given the paired feature maps $\{F^{M}|_{M=R,T}\} \in \mathbb{R}^{c \times h \times w}$, we first introduce three different linear layers to learn three new feature representations, denoted by $\{Q^{M},K^{M},V^{M}\}\in \mathbb{R}^{c \times h \times w}$. Note that we omit the subscript $l$ denoting the layer for presentation clarity in this subsection. Based on $Q^{M}$, $K^{M}$, and $V^{M}$, the CRP and SRP blocks perform relation-propagation for capturing modality-shared features.

\subsubsection{Channel relation-propagation}
Taken as input $Q^{M}$, $K^{M}$, and $V^{M}$, the CRP block first calculates the inter-channel relation matrix $W^{M}_{ch}\in \mathbb{R}^{c\times c}$ for every modality. Denoting the vectorial representation of the $m$-th channel of $Q^{M}$ and the $n$-th channel of $K^{M}$ as $q^{M}_{ch,m}$ and $k^{M}_{ch,n}$, respectively, the $(m,n)$-th weight value in $W^{M}_{ch}$ is computed via a relation function $f(\cdot,\cdot)$, which can be formulated as:
\begin{equation}
\label{Eq:relation_function}
w^{M}_{ch,(m,n)}=f(q^{M}_{ch,m}, k^{M}_{ch,n}).
\end{equation}

Herein we consider two choices for $f(\cdot,\cdot)$: dot product and Gaussian function.

\noindent\textbf{Dot product.}
Similar to~\cite{non-local,transformer}, the relation function $f(\cdot,\cdot)$ can be defined as the dot product:
\begin{equation}
\label{Eq:dot_product}
f(x, y) = x\cdot y,
\end{equation}
in which $\cdot$ denotes the dot product operation.

\noindent\textbf{Gaussian function.}
Following the non-local network~\cite{non-local}, a natural choice of $f$ is the Gaussian function, which can be formulated as:
\begin{equation}
\label{Eq:gaussian_function}
f(x, y) = {\rm exp}(x\cdot y).
\end{equation}

Different feature channels represent various semantics. Thus, the calculated inter-channel relation matrix models the long-range semantic dependencies between different feature channels. A large relation weight denotes the strong semantic dependency between the corresponding channels. With $W_{ch}^{R}$ and $W_{ch}^{T}$, we propagate the semantic relation between different modalities via performing the element-wise product between $W_{ch}^{R}$ and $W_{ch}^{T}$, which enables us to identify the shared semantic dependencies between the RGB and thermal features. Formally, the shared relation matrix $P_{ch}\in \mathbb{R}^{c \times c}$ can be formulated as:
\begin{equation}
\label{Eq:channel_relation}
 P_{ch}=\sigma_{n}({\rm ReLU}(W^{R}_{ch})\odot{\rm ReLU}(W^{T}_{ch})).
\end{equation}
Herein $\odot$ refers to the element-wise product operation, and $\sigma_{n}$ denotes the normalization operation used to normalize the sum of each row of the input matrix to 1. The ReLU function in Eq.~\ref{Eq:channel_relation} is used to filter out the negative relation weight, which can be omitted when using the Gaussian function as the relation function $f$. 
After that, we use the shared relation matrix $P_{ch}$ as the attention weights to capture the modality-shared features, which can be implemented via matrix multiplication:
\begin{equation}
\begin{split}
&S^{R}_{ch}= P_{ch}V^{R},\\
&S^{T}_{ch}= P_{ch}V^{T}.
\end{split}
\end{equation}
Herein $V^{M}$ is reshaped to $c\times hw$ before performing matrix multiplication, and $S^{R}_{ch}$ and $S^{T}_{ch}$ denote the captured modality-shared features for the RGB and thermal images, respectively.

\subsubsection{Spatial relation-propagation} 
Similar to the CRP block, taken as input $Q^{M}$, $K^{M}$, and $V^{M}$, the SRP block first computes the inter-pixel relation matrix $W^{M}_{sp}\in \mathbb{R}^{hw\times hw}$ for every modality.
Denoting the vectorial representation of the $u$-th pixel of $Q^{M}$ and the $v$-th pixel of $K^{M}$ as $q^{M}_{sp,u}$ and $k^{M}_{sp,v}$, respectively, the $(u,v)$-th weight in $W^{M}_{sp}$ can be computed via the relation function defined by Eq.~\ref{Eq:dot_product} or Eq.~\ref{Eq:gaussian_function}. With the calculated inter-pixel relation matrices $W^{R}_{sp}$ and $W^{T}_{sp}$, we perform element-wise product between them to obtain the shared relation matrix $P_{sp}\in \mathbb{R}^{hw \times hw}$:
\begin{equation}
\label{Eq:spatial_relation}
 P_{sp}=\sigma_{n}({\rm ReLU}(W^{R}_{sp})\odot{\rm ReLU}(W^{T}_{sp})).
\end{equation}
After that, we perform matrix multiplication to capture the features containing the modality-shared spatial information $S^{R}_{sp}$ and $S^{T}_{sp}$:
\begin{equation}
\label{Eq:channel_rgb_information}
\begin{split}
 & S^{R}_{sp} = V^{R} P_{sp},\\
 & S^{T}_{sp} = V^{T} P_{sp}.
\end{split}
\end{equation}

\subsubsection{Interactive feature fusion}
With the captured modality-shared features $S^{R}_{ch}$, $S^{T}_{ch}$, $S^{R}_{sp}$, and $S^{T}_{sp}$, we aggregate them with the input features via the element-wise summation operation to enhance the input features. To be more specific, the modality-shared feature extracted from one modality will be aggregated with the input feature of the other modality. For example, the modality-shared features extracted from the RGB image $S^{R}_{ch}$ and $S^{R}_{sp}$ will be aggregated with the input feature of the thermal modality $F^{T}$. In this way, we can exploit the modality-shared features of one modality to enhance the features of the other modality without contaminating its modality-specific information. The above aggregation process can be formulated as:
\begin{equation}
\label{Eq:fuse_feature}
\begin{split}
& F^{R}_{enhance}=F^{R} \oplus {\lambda}^{T}_{ch}S^{T}_{ch} \oplus {\lambda}^{T}_{sp}S^{T}_{sp},\\
& F^{T}_{enhance}=F^{T} \oplus {\lambda}^{R}_{ch}S^{R}_{ch} \oplus {\lambda}^{R}_{sp}S^{R}_{sp}.
\end{split}
\end{equation}
Herein $\oplus$ denotes the element-wise summation operation. ${\lambda}^{R}_{ch}$, ${\lambda}^{R}_{sp}$, ${\lambda}^{T}_{ch}$, and ${\lambda}^{T}_{sp}$ are learnable parameters, whose initialization values are set to 1 at beginning of the training in our experiments. While feeding $F^{R}_{enhance}$ and $F^{T}_{enhance}$ back to the feature extraction block of the next layer, we also fuse them together to obtain the fused feature $F^{C}$.

\begin{figure*}[t]
\centering
\includegraphics[width=1.0\textwidth]{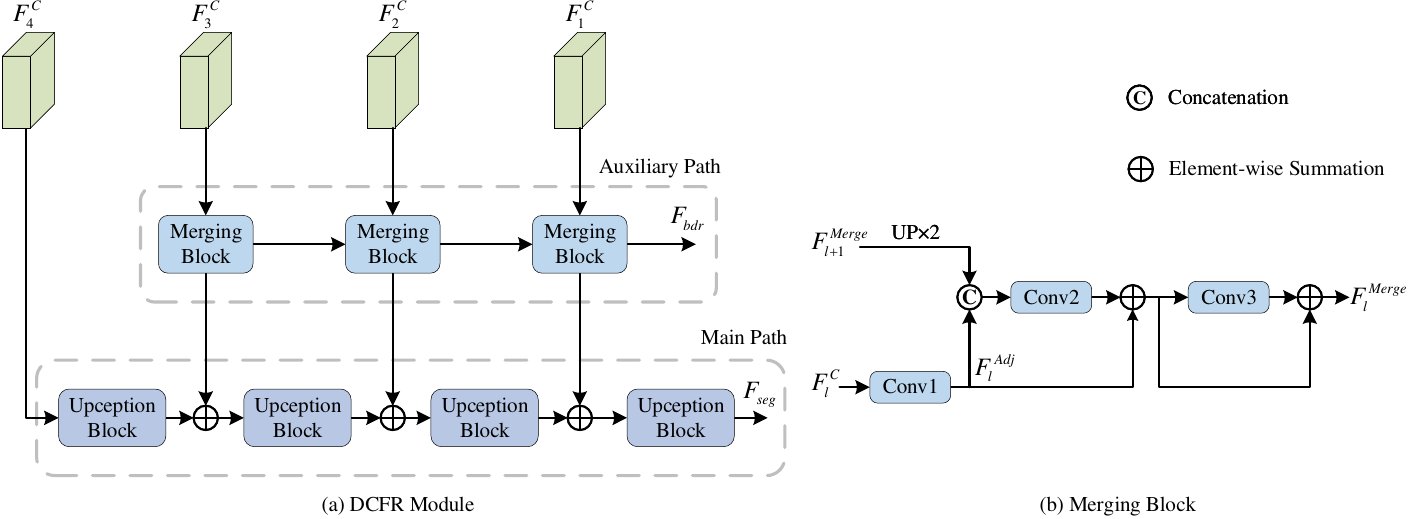}
\caption{\textbf{Structure of our proposed DCFR module and the merging block.} (a) The DCFR module consists of two feature refinement paths: the auxiliary path and the main path, which output two refined features for boundary prediction and semantic prediction, respectively. (b) The merging block refines the feature from a deeper layer with more semantics using the feature from a shallower layer with more spatial details.}
\label{Fig:DCFR}
\end{figure*}

\subsection{Dual-path cascaded feature refinement}
As above-mentioned, we introduce boundary prediction as an auxiliary task into our architecture to improve the ability of our CSRPNet to perceive the boundary between different semantic regions, which has been proven effective by~\cite{EGFNet,swinnet}. Unlike the semantic segmentation task, the boundary prediction task only necessitates the model to predict whether a pixel falls on a boundary between different semantic regions. Considering this difference, we propose a Dual-path Cascaded Feature Refinement (DCFR) module to generate two different refined feature maps for boundary prediction and semantic segmentation, respectively. Besides, the cascaded refinement mechanism in the DCFR module enables our method to harness the potential of the multi-layer features. As shown in Figure~\ref{Fig:DCFR} (a), the DCFR module consists of an auxiliary path and a main path, taking as input the multi-layer fused features $\{F^{C}_{l}|_{l=1,2,3,4}\}$ and outputting two refined feature maps.

The auxiliary path is mainly composed of cascaded merging blocks. Every merging block, except the first one, accepts the fused feature $F^{C}_{l}$ and the output feature of the last merging block $F^{Merge}_{l+1}$ as inputs. In the inputs, $F^{C}_{l}$ with higher resolution contains more spatial details, and $F^{Merge}_{l+1}$ from a deeper layer contains more semantic information. Merging them yields more informative feature representations. Figure~\ref{Fig:DCFR} (b) illustrates the structure of the merging block. Similar to~\cite{priorNet}, it first uses a $1\times 1$ convolutional layer to reduce the channel numbers of $F^{C}_{l}$, yielding the channel-downsampled feature $F^{Adj}_{l}$. Meanwhile, it uses spatial up-sampling to adjust the spatial size of $F_{l+1}^{Merge}$ to that of $F^{Adj}_{l}$. Then it concatenates $F^{Adj}_{l}$ and the up-sampled $F_{l+1}^{Merge}$ and processes the concatenated feature using a $1\times 1$ convolutional layer with a skip connection. After that, we use a $3\times 3$ convolutional layer with a skip connection to generate the final merged feature. This process can be formulated as:
\begin{equation}
\label{Eq:merging_block}
\begin{split}
& F_{l}^{Adj} = \phi_{conv1}(F_{l}^{C}),\\
& F_{l}^{E} = \phi_{conv2}(F_{l}^{Adj}\uplus \phi_{up}(F_{l+1}^{Merge})) \oplus F_{l}^{Adj},\\
& F_{l}^{Merge} = {\rm ReLU}(\phi_{conv3}(F_{l}^{E})) \oplus F_{l}^{E},
\end{split}
\end{equation}
where $\phi_{conv1}$, $\phi_{conv2}$, and $\phi_{conv3}$ denote the three convolutional layers in Figure~\ref{Fig:DCFR} (b). $\phi_{up}$ denotes the spatial up-sampling operation. $\uplus$ and $\oplus$ denote the concatenation and element-wise summation operations, respectively.
In the auxiliary path, we only use the shallow fused features $\{F^{C}_{l}|l=3,2,1\}$, since predicting boundary maps requires more spatial details than semantic information. For the first merging block that only accepts the fused feature as input, we remove the up-sample and concatenation operations.

In the main path, we take the fused features from all layers into account as both the semantic information and spatial details are crucial for semantic segmentation. The main path is composed of the cascaded Upception blocks~\cite{RTFNet}, which are used to increase the spatial resolution of the input features. In addition, we introduce the skip connections between the auxiliary path and the main path. This design enables the main path to exploit the boundary information learned by the auxiliary path from the boundary prediction task.

\subsection{Classifiers}
On top of the output features $F_{bdr}$ and $F_{seg}$, we construct a boundary classifier and a semantic classifier to predict the boundary map and the semantic map, respectively. Specifically, the boundary classifier is composed  of a $1 \times 1$ convolutional layer and a $3 \times 3$ convolutional layer, following the design of~\cite{EGFNet}. The semantic classifier is constructed using an Upception block~\cite{RTFNet}.

\subsection{Loss function}
To train our CSRPNet, we adopt a multi-task loss function that imposes supervision on the predicted boundary map and semantic map. The multi-task loss function can be formulated as:
\begin{equation}
\label{Eq:RGB_fused_feature}
    \mathcal{L}_{total}={\lambda}_{1} \mathcal{L}_{bdr} + {\lambda}_{2} \mathcal{L}_{seg},
\end{equation}
where $\mathcal{L}_{bdr}$ and $\mathcal{L}_{seg}$ denote the boundary loss and the semantic segmentation loss, respectively, and ${\lambda}_{1}$ and ${\lambda}_{2}$ are the balance weights. We utilize a sliding window mechanism to obtain the boundary label. We define the center pixel of the sliding window as a boundary pixel if there is more than one semantic class in the sliding window. Herein the size of the sliding window is set to $5 \times 5$.
Considering the imbalance between the boundary pixels and the non-boundary pixels, we use the weighted cross-entropy loss as our boundary loss, in which the weights are set following~\cite{Enet}. In addition, we use the cross-entropy loss as our semantic segmentation loss.

%% file: experiments.tex
\subsection{Implementation details}
We implement the proposed CSRPNet with the PyTorch toolkit on a computer equipped with the NVIDIA GTX 2080Ti GPU. We use a two-stage learning strategy for training. In the first stage, we only optimize the learnable parameters in the encoder, the main path of the DCFR module, and the semantic classifier. In the second stage, we freeze the parameters optimized in the first stage and optimize the learnable parameters in the auxiliary path of the DCFR module and the boundary classifier.
In particular, we use the Stochastic Gradient Descent (SGD) method with a momentum of 0.9 and a weight decay of 0.0005 as the optimizer in both two training stages. The initial learning rates in both two training stages are set to 0.01 and decrease in an exponential manner with a decay factor of 0.95. The balance weights $\lambda_1$ and $\lambda_2$ are tuned to be both 1. In addition, the data augmentation methods including random flipping and random cropping are used to alleviate the problem of insufficient training data.

\begin{table}[t]
\centering
\rowcolors{2}{white}{gray!25}
\setlength{\tabcolsep}{5pt}
\renewcommand{\arraystretch}{1.5}
\caption{\textbf{Experimental results of the six variants of our method on the MFNet dataset.} Herein dot product is used as the relation function. The best and second-best results are denoted by the bold font and the underline, respectively.}
\resizebox{0.65\linewidth}{!}{
\begin{tabular}{lccccccccc}
\toprule
 \multirow{2}{*}{Variants}   &  \multicolumn{2}{c}{Overall} && \multicolumn{2}{c}{Daytime} && \multicolumn{2}{c}{Nighttime}  \\
    \cmidrule{2-3} \cmidrule{5-6} \cmidrule{8-9} 
 ~  & mAcc & mIoU && mAcc & mIoU && mAcc & mIoU\\
\midrule
(a) Baseline & 59.1 & 48.4 && 55.8 & 43.4 && 56.0 & 48.5 \\
(b) +SRP  & 65.6 & 53.1 && 62.8 & 46.4 && 61.6 & 53.4 \\
(c) +CRP  & 67.6 & 53.3 && 61.2 & 45.6 && \underline{64.8} & 54.7 \\
(d) +CSRP    & \underline{67.7} & 54.1 && \underline{64.2} & 47.3 && 62.9 & 54.0 \\
(e) +CSRP+DCFR (Ours) & \textbf{70.4} & \textbf{55.5} && \textbf{65.5} & \textbf{48.0} && \textbf{66.0} & \textbf{55.6} \\
(f) Ours w/o Boundary & \underline{67.7} & \underline{55.2} && 61.8 & \underline{47.9} && 63.5 & \underline{55.2}\\
\bottomrule
\end{tabular}}
\label{Table: ablation}
\end{table}

\begin{figure}[t]
\centering
\includegraphics[width=1.0\columnwidth]{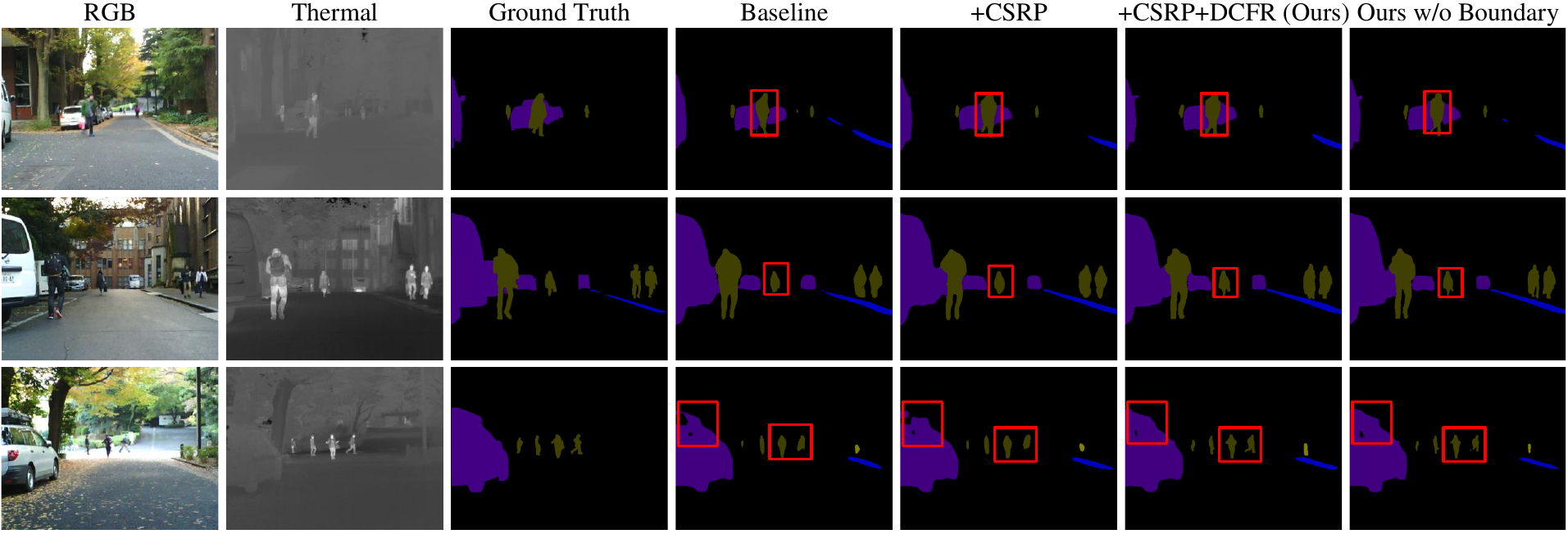} 
\caption{\textbf{Visual comparisons between different variants.} We use the red boxes to denote the parts where the segmentation results are different between different variants. The comparison results show that our intact model with both the CSRP and DCFR modules achieves the best segmentation performance.}
\label{Fig: ablation_visualization}
\end{figure}

\subsection{Benchmarks and metrics}
We evaluate CSRPNet on two popular benchmarks, including MFNet~\cite{MFNet} and PST900~\cite{PSTNet}. The MFNet dataset contains nine semantic classes, including \emph{Car}, \emph{Person}, \emph{Bike}, \emph{Curve}, \emph{Car stop}, \emph{Guardrail}, \emph{Color cone}, and \emph{Bump}. It consists of 1569 annotated RGB and thermal image pairs in total, in which 820 and 749 image pairs are taken in the daytime and nighttime, respectively. 
All images are resized to $480 \times 640$. For a fair comparison, we use the same training and evaluation configurations as those in MFNet~\cite{MFNet}. In particular, we follow MFNet~\cite{MFNet} to split the data into the training, validation, and testing sets, which contain 784, 392, and 393 image pairs, respectively. The PST900~\cite{PSTNet} dataset is composed of 894 aligned RGB and thermal image pairs and contains four visible artifacts, including \emph{Fire-Extinguisher}, \emph{Backpack}, \emph{Hand-Drill}, and \emph{Survivor}.
Following MFNet~\cite{MFNet}, we use the widely used mean accuracy (mAcc) and mean intersection over union (mIoU) to quantitatively evaluate segmentation performance. The mean accuracy metric is also known as recall. Formally, these two metrics are defined as:
\begin{equation}
\label{Eq:metrics}
\begin{split}
&\mathrm{mAcc}\!=\!\frac{1}{N}\! \sum_{i=1}^{N} \frac{\sum_{k=1}^{K} p_{i i}^{k}}{\sum_{k=1}^{K} p_{i i}^{k}\!+\!\sum_{k=1}^{K} \sum_{j=1, j \neq i}^{N} p_{i j}^{k}}, \\
&\mathrm{mIoU}\!=\!\frac{1}{N}\! \sum_{i=1}^{N} \frac{\sum_{k=1}^{K} p_{i i}^{k}}{\sum_{k=1}^{K} p_{i i}^{k}\!+\!\sum_{k=1}^{K} \sum_{j=1, j \neq i}^{N}(p_{j i}^{k}\!+\!p_{i j}^{k})}.
\end{split}
\end{equation}
Herein $N$ is the number of semantic classes. $K$ is the number of RGB-T image pairs. $p_{i i}^{k}$ is the number of pixels that belong to class $i$ and are correctly classified as class $i$ in the $k$-th RGB-T image pair. $p_{j i}^{k}$ is the number of pixels that belong to class $j$ but are incorrectly classified as class $i$ in the $k$-th RGB-T image pair. $p_{i j}^{k}$ is the number of pixels that belong to class $i$ but are incorrectly classified as class $j$ in the $k$-the RGB-T image pair. 

\begin{table}[t]
\centering
\rowcolors{2}{white}{gray!25}
\setlength{\tabcolsep}{5pt}
\renewcommand{\arraystretch}{1.5}
\caption{\textbf{Experimental results of deploying the proposed CSRP module in different layers on the MFNet dataset.} The notation $\checkmark$ represents that we deploy the proposed CSRP module in the corresponding layer.}
\resizebox{0.65\linewidth}{!}{
\begin{tabular}{ccccccccccc}
\toprule
\multirow{2}{*}{Layer2} & \multirow{2}{*}{Layer3} & \multirow{2}{*}{Layer4} & \multicolumn{2}{c}{Overall} && \multicolumn{2}{c}{Daytime} && \multicolumn{2}{c}{Nighttime}\\
\cmidrule{4-5} \cmidrule{7-8} \cmidrule{10-11}
 &&& mAcc & mIoU && mAcc & mIoU && mAcc & mIoU\\
\midrule
~ & ~  & ~ & 59.1 & 48.4 && 55.8 & 43.4 && 56.0 & 48.5 \\
\checkmark & ~  & ~ & 60.1 & 50.5 && 58.1 & 45.0 && 56.7 & 50.2 \\
~ & \checkmark & ~ & 62.9 & 53.1 && 59.5 & \underline{46.9} && 58.5 & 51.9 \\
~ & ~ & \checkmark & 65.2 & 53.1 && 60.9 & \underline{46.9} && 60.8 & 52.5\\
\checkmark & \checkmark & ~ & 63.8 & 51.7 && 60.3 & 45.7 && 59.7 & 51.6\\
\checkmark & ~ & \checkmark & 62.6 & 50.6 && 58.4 & 44.1 && 59.3 & 51.6\\
~ & \checkmark & \checkmark & \underline{65.8} & \textbf{54.3} && \underline{61.5} & \underline{46.9} && \underline{62.3} & \textbf{54.6}\\
\checkmark & \checkmark & \checkmark & \textbf{67.7} & \underline{54.1} && \textbf{64.2} & \textbf{47.3} && \textbf{62.9} & \underline{54.0}\\
\bottomrule
\end{tabular}}
\label{Table: ablation1}
\end{table}
\subsection{Ablation study}
We first conduct experiments to investigate the effect of each proposed component and the effect of using features from different layers. Note that we use the dot product as the relation function in this section.

\begin{table}[t]
\centering
\rowcolors{2}{gray!25}{white}
\setlength{\tabcolsep}{1.9pt}
\renewcommand{\arraystretch}{1.5}
\caption{\textbf{Experimental results of different algorithms on MFNet dataset.} `3c' and `4c' denote that the corresponding model is tested with the three-channel RGB data and the four-channel RGB-T data, respectively. Note that mAcc and mIoU are calculated with the unlabeled classes while the results for the unlabeled classes are not displayed.}
\resizebox{1.0\linewidth}{!}{
\begin{tabular}{lc ccccccccccccccccccccccccc cc}
\toprule
\multirow{2}{*}{Methods} & \multirow{2}{*}{Backbone} && \multicolumn{2}{c}{Car} && \multicolumn{2}{c}{Person} && \multicolumn{2}{c}{Bike} && \multicolumn{2}{c}{Curve} && \multicolumn{2}{c}{Car stop} && \multicolumn{2}{c}{Guardrail} && \multicolumn{2}{c}{Color cone} && \multicolumn{2}{c}{Bump} &&  \multirow{2}{*}{mAcc} & \multirow{2}{*}{mIoU}\\
\cmidrule{4-5}  \cmidrule{7-8}  \cmidrule{10-11}  \cmidrule{13-14}  \cmidrule{16-17}  \cmidrule{19-20}  \cmidrule{22-23} \cmidrule{25-26}
~ & ~ &&Acc & IoU && Acc & IoU && Acc & IoU && Acc & IoU && Acc & IoU && Acc & IoU && Acc & IoU && Acc & IoU  && ~ & ~  \\
\midrule
FRRN (3c) & ResNet && 80.0 & 71.2 && 53.0 & 46.1 && 65.1 & 53.0 && 34.0 & 27.1 && 21.6 & 19.1 && 0.0 & 0.0 && 34.7 & 32.5 && 36.2 & 30.5 && 47.1 & 41.8\\
FRRN (4c) & ResNet && 81.9 & 74.7 && 66.2 & 60.8 && 62.8 & 50.3 && 41.2 & 35.0
&& 12.5 & 11.5 && 0.0 & 0.0 && 37.2 & 34.0 && 35.2 & 34.6 && 48.5 & 44.2\\
DFN (3c) & ResNet && 90.7 & 81.4 && 67.7 & 52.8 && 71.5 & 57.5 && 49.2 & 34.9
&& 35.1 & 23.8 && 4.1 & 1.4 && 44.2 & 31.0 && 54.6 & 47.5 && 57.3 & 47.5\\
DFN (4c) & ResNet && 90.0 & 84.4 && 73.2 & 65.0 && 75.5 & 60.9 && 54.0 & 40.4
&& 38.9 & 25.7 && 10.2 & 2.7 && 48.3 & 42.5 && 55.8 & 47.4 && 60.5 & 52.0\\
BiSeNet (3c) & U-Net && 90.0 & 84.5 && 65.0 & 54.3 && 75.0 & 61.4 && 32.1 & 25.7 && 32.3 & 26.2 && 3.2 & 0.9 && 49.6 & 43.3 && 48.1 & 40.5 && 54.9 & 48.2\\
BiSeNet (4c) & U-Net && 89.7 & 84.1 && 72.0 & 63.2 && 74.1 & 60.1 && 45.1 & 36.7 && 34.2 & 25.3 && 18.2 & 5.0 && 47.4 & 42.2 && 39.8 & 35.9 && 57.7 & 50.0\\
MFNet & CNN && 77.2 & 65.9 && 67.0 & 58.9 && 53.9 & 42.9 && 36.2 & 29.9
&& 12.5 & 9.9 && 0.1 & 0.0 && 30.3 & 25.2 && 30.0 & 27.7 && 45.1 & 39.7\\
FuseNet & VGG && 81.0 & 75.6 && 75.2 & 66.3 && 64.5 & 51.9 && 51.0 & 37.8
&& 28.7 & 15.0 && 0.0 & 0.0 && 31.1 & 21.4 && 51.9 & 45.0 && 52.4 & 45.6\\
DepthAwareCNN & CNN && 85.2 & 77.0 && 61.7 & 53.4 && 76.0 & 56.5 && 40.2
& 30.9 && 9.9 & 29.3 && 22.8 & 6.4 && 32.9 & 30.1 && 36.5 & 32.3 && 55.1 & 46.1\\
RTFNet-50 & ResNet && 91.3 & 86.3 && 78.2 & 67.8 && 71.5 & 58.2 && \underline{69.8}
& 43.7 && 32.1 & 24.3 && 13.4 & 3.6 && 40.4 & 26.0 && 73.5 & \textbf{57.2} && 62.2
& 51.7\\
RTFNet-152 & ResNet && 93.0 & 87.4 && 79.3 & 70.3 && 76.8 & \underline{62.7} && 60.7
& 45.3 && 38.5 & 29.8 && 0.0 & 0.0 && 45.5 & 29.1 && \underline{74.7} & 55.7 && 63.1
& 53.2\\
MMNet & ResNet && -- & 83.9 && -- & 69.3 && -- & 59.0 && -- & 43.2
&& -- & 24.7 && -- & 4.6 && -- & 42.2 && -- & 50.7 && 62.7 & 52.8\\
FuseSeg-161 & DenseNet && 93.1 & \textbf{87.9} && 81.4 & 71.7 && 78.5 & \textbf{64.6}
&& 68.4 & 44.8 && 29.1 & 22.7 && \underline{63.7} & 6.4 && 55.8 & 46.9 && 66.4 & 47.9
&& 70.6 & 54.5\\
ABMDRNet & ResNet && \underline{94.3} & 84.8 && \textbf{90.0} & 69.6 && 75.7 & 60.3 && 64.0
& 45.1 && 44.1 & 33.1 && 31.0 & 5.1 && 61.7 & 47.4 && 66.2 & 50.0 && 69.5 & 54.8\\
EGFNet & ResNet && \textbf{95.8} & 87.6 && 89.0 & 69.8 && \textbf{80.6} & 58.8 && \textbf{71.5} & 42.8
&& \underline{48.7} &  \underline{33.8} && 33.6 & 7.0 && \underline{65.3} &48.3 && 71.1 & 47.1 && \underline{72.7} & 54.8\\
GCNet & ResNet && 94.2 & 86.0 && \underline{89.6} & 72.0 && 77.5 & 60.0 && 68.9 & 42.8
&& 38.3 & 30.7 && 45.8 & 6.2 && 59.6 & \underline{49.5} && \textbf{82.1} & 52.6 && \underline{72.7} & 55.3\\
FEANet & ResNet && 93.3 & \underline{87.8} && 82.7 & 71.1 && 76.7 & 61.1 && 65.5 & 46.5
&& 26.6 & 22.1 && \textbf{70.8} & 6.6 && \textbf{66.6} & \textbf{55.3} && 77.3 & 48.9 && \textbf{73.2} & 55.3\\
GCGLNet & MobileV2 && 92.4 & 83.8 && 85.7 & \textbf{72.4} && 73.7 &59.6 && 68.1 & 40.9
&& \textbf{54.9} & \textbf{43.0} && 52.3 & \textbf{8.7} && 48.8 & 43.8 && 73.2 & 48.6 && 72.0 & 55.4\\
MMDRNet & ResNet && 93.4 & 85.7 && 89.3 & 70.3 && 74.7 & 61.5 && 65.7 & \underline{46.9}
&& 42.7 & 32.7 && 53.9 & 7.7 && 59.9 & 48.2 && 73.0 & 53.4 && 72.4 & \textbf{56.0}\\
\textbf{CSRPNet} (Dot product) & ResNet && 93.7 & 87.6 && 86.5 & \underline{72.3} && 78.1 & 60.7 && \underline{69.8} & 45.0
&& 40.2 & 29.8 && 36.1 & 6.4 && 58.5 & 46.3 && 71.5 & 53.2 && 70.4 & \underline{55.5}\\
\textbf{CSRPNet} (Gaussian) & ResNet && 93.5 & \underline{87.8} && 87.8 & \underline{72.3} && \underline{78.9} & 61.4 && 68.5 & \textbf{47.6} && 38.9 & 28.1 && 58.9 & \underline{8.0} && 57.3 & 43.7 && 71.2 & \underline{57.1} && \underline{72.7} & \textbf{56.0}\\
\bottomrule
\end{tabular}}
\label{Table: compare}
\end{table}

\newcommand{\tabincell}[2]{\begin{tabular}{@{}#1@{}}#2\end{tabular}}
\begin{table}
\centering
\setlength{\tabcolsep}{2.5pt}
\renewcommand{\arraystretch}{1.5}
\caption{\textbf{Experimental results on the MFNet dataset for the daytime and nighttime images.}}
\resizebox{1.0\linewidth}{!}{
\begin{tabular}{lc cccccccccccccc cc}
\toprule
\multicolumn{2}{l}{Methods} & \tabincell{c}{FRRN\\(3c)} & \tabincell{c}{FRRN\\(4c)} & \tabincell{c}{BiSeNet
\\(3c)} & \tabincell{c}{BiSeNet\\(4c)} & \tabincell{c}{DFN\\(3c)} & \tabincell{c}{DFN\\(4c)} & \tabincell{c}{Fuse\\Net}&\tabincell{c}{ABMDR\\Net}&\tabincell{c}{Fuse\\Seg-161}&\tabincell{c}{Depth\\AwareCNN}& \tabincell{c}{MFNet} & \tabincell{c}{RTFNet\\-50} & \tabincell{c}{RTFNet\\-152} & \tabincell{c}{\textbf{CSRPNet}\\(Dot product)}  &\tabincell{c}{\textbf{CSRPNet}\\(Gaussian)} \\
\midrule
\multirow{2}{*}{Daytime} & mAcc & 45.1 & 42.4 & 52.1 & 52.9 & 53.7 & 53.4 & 49.5 &58.4&62.1& 50.6& 42.6 & 57.3 & 60.0 & \textbf{65.5} & \underline{64.0} \\
& mIoU & 40.0 & 38.0 & 44.5 & 44.8 & 42.2 & 43.9 & 41.0&46.7&47.8& 42.4& 36.1 & 44.4 & 45.8 & \underline{48.0} & \textbf{48.7}  \\
\midrule
\multirow{2}{*}{Nighttime} & mAcc & 41.6 & 46.2 & 50.3 & 53.1 & 52.4 & 57.4 & 48.9&\underline{68.3}&67.3& 50.7& 41.4 & 59.4 & 60.7 & 66.0 & \textbf{69.4} \\
& mIoU & 37.3  & 42.3 & 45.0 & 47.7 & 44.6 & 51.8 & 43.9&55.5&54.6& 43.2& 36.8 & 52.0 & 54.8 & \underline{55.6} & \textbf{56.0}  \\
\bottomrule
\end{tabular}}
\label{Table: day_night}
\end{table}

\begin{table}[t]
\centering
\rowcolors{2}{white}{gray!25}
\renewcommand{\arraystretch}{1.5}
\caption{\textbf{Experimental results of different algorithms on PST900.}}
\resizebox{1.0\linewidth}{!}{
\begin{tabular}{lccccccccccccccccccc}
\toprule
\multirow{2}{*}{Methods} &\multirow{2}{*}{Backbone} & \multicolumn{2}{c}{Background} && \multicolumn{2}{c}{Hand-Drill} && \multicolumn{2}{c}{Backpack} && \multicolumn{2}{c}{Fire-extinguisher} && \multicolumn{2}{c}{Survior} &&  \multirow{2}{*}{mAcc} & \multirow{2}{*}{mIoU}\\
\cmidrule{3-4}\cmidrule{6-7}\cmidrule{9-10}\cmidrule{12-13}\cmidrule{15-16}
 & &Acc & IoU && Acc & IoU && Acc & IoU && Acc & IoU && Acc & IoU &&  &   \\
\midrule
Efficient FCN (3c)&VGG-16& \underline{99.8} & 98.6 &&32.1 & 30.2 && 60.1 & 58.2&& 78.9 & 40.0 && 32.8 & 28.0 && 60.7 & 51.0\\
Efficient FCN (4c)&VGG-16& \underline{99.8} & 98.9 &&48.8 & 38.6 && 69.9 & 67.6&& 76.5 & 46.3 && 38.9 & 35.1 && 66.8 & 57.3\\
MFNet&DCNN & \textbf{99.9} & 98.7 &&46.7 & 39.3 && 52.8 & 52.4&& 71.8 & 67.4 && 18.8 & 18.9 && 58.0 & 55.3\\
RTFNet&ResNet-152 & \underline{99.8} & 99.0 &&7.8 & 7.1 && 80.0 & 74.2&& 62.4 & 51.9 && \textbf{78.5} & \textbf{70.1} && 65.7 & 60.5\\
CCNet (3c)&ResNet-50 & \textbf{99.9} & 99.1 &&51.8 & 32.3 && 68.3 & 66.4&& 67.8 & 51.8 && 60.8 & 57.5 && 69.7 & 61.4\\
CCNet (4c)&ResNet-50 & 99.6 & 98.7 &&54.1 & 51.0 && 76.0 & 73.0&& \textbf{88.1} & \textbf{73.8} && 49.5 & 33.5 && 73.4 & 66.0\\
PSTNet&ResNet-18 & -- & 98.9 &&-- &53.6 && -- & 69.2&& -- & \underline{70.1} && -- & 50.0 && -- & 68.4\\
ACNet&ResNet-50 & \underline{99.8} & \textbf{99.3} &&53.6 & 51.7 && 85.6 & \textbf{83.2}&& 84.9 & 60.0 &&\underline{ 69.1} & \underline{65.2} && 78.7 & 71.8\\
\textbf{CSRPNet} (Dot product)&ResNet-50 & \underline{99.8} & \textbf{99.3} && \underline{73.7} & \underline{63.2} && \textbf{87.3} & \underline{82.6} &&\underline{86.3} & 67.7 && 60.0 & 56.6 && \textbf{81.4} & \underline{73.9}\\
\textbf{CSRPNet} (Gaussian)&ResNet-50 & \underline{99.8} & \underline{99.2} && \textbf{86.1} & \textbf{71.6} && \underline{86.8} & 82.0 &&77.7 & 67.9 && 55.2 & 54.1 && \underline{81.1} & \textbf{75.0}\\
\bottomrule
\end{tabular}}
\label{Table: compare}
\end{table}

\subsubsection{Effect of each component}
To analyze the effect of each component, we evaluate six variants of our algorithm: (a) the baseline (BL) model without the proposed CSRP and DCFR modules, which directly fuses the features of the RGB and thermal images from the last layer via element-wise summation and uses the decoder in MFNet~\cite{MFNet} to predict the semantic map; (b) the variant that exploits the SRP block for multi-modality feature fusion based on the baseline model; (c) the variant that uses the CRP block for feature fusion based on the baseline model; (d) the variant that uses the proposed CSRP module for feature fusion based on the baseline model; (e) our intact model including the proposed CSRP and DCFR modules; (f) the variant that removes the boundary prediction branch and the boundary supervision from our intact model. Table~\ref{Table: ablation} reports the overall, daytime, and nighttime performance of these variants on the MFNet datasets.

Compared with the baseline model, the variants using the SRP and CRP blocks achieve performance gains of 6.5\%/8.5\% and 4.7\%/4.9\% in overall mAcc and mIoU, respectively. These performance gains demonstrate the effectiveness of spatial or channel relation-propagation. Compared with the variants using the SRP block or the CRP block alone, exploiting them together (the variant using the CSRP module) further promotes segmentation performance, which manifests that the relation-propagation in space and channel is complementary. Furthermore, the performance gap in overall mAcc between the variant using the CSRP module and our intact model validates the effectiveness of the DCFR module. Besides, removing the boundary prediction branch and the boundary supervision leads to performance drops, demonstrating their effectiveness. The comparisons between these variants in terms of daytime and nighttime performance also demonstrate the effectiveness of each proposed component.

We also visualize the segmentation results of different variants to analyze the effect of the CSRP and DCFR modules, as shown in Figure~\ref{Fig: ablation_visualization}. Compared with the baseline model, the variant using the CSRP module predicts more accurate segmentation masks. The comparison between the last two rows shows that the DCFR module enables our method to predict more refined masks, especially for small objects. Removing the boundary prediction branch and boundary supervision leads to a less precise segmentation.

\subsubsection{Effect of using features from different layers} We also conduct experiments to analyze the effect of deploying the CSRP module in different layers of the encoder based on the above-mentioned variant (d).
Table~\ref{Table: ablation1} reports the detailed experimental results. Considering that calculating the relation matrix of the feature map of the first layer with a high resolution is memory-consuming and the low-level features lack abstract semantic information, we omit the CSRP module in the first layer of the encoder. From the first to the fourth rows in Table~\ref{Table: ablation1}, we can observe that deploying the proposed CSRP module in the deeper layer leads to better performance, which demonstrates that the feature from the deeper layer models more abstract semantics to extract the modality-shared information. In addition, we can observe that deploying the proposed CSRP module in \emph{layer2}, \emph{layer3}, and \emph{layer4} leads to the best segmentation performance. 

\subsection{Comparison with state-of-the-art methods}
In this section, we evaluate our method on the MFNet~\cite{MFNet} and PST900~\cite{PSTNet} datasets. We detail the comparison results per dataset in the following.

\vspace{1mm}
\noindent\textbf{The MFNet dataset.}
We compare our proposed CSRPNet with 15 state-of-the-art methods, including FRRN~\cite{FRRN}, BiSeNet~\cite{bisenet}, DFN~\cite{DFN}, FuseNet~\cite{Fusenet}, DepthAwareCNN~\cite{DepthAwareCNN}, MFNet~\cite{MFNet}, RTFNet~\cite{RTFNet}, FuseSeg-161~\cite{FuseSeg}, ABMDRNet~\cite{ABMDRNet}, EGFNet~\cite{EGFNet}, GCNet~\cite{GCNet}, MMDRNet~\cite{MMDRNet}, GCGLNet~\cite{GCGLNet}, FEANet~\cite{FEANet}, MMNet~\cite{MMNet}, on the MFNet dataset. 
Table~\ref{Table: compare} reports the experimental results. For a fair comparison, we test two versions of several RGB segmentation methods. One is the `3c' version taking as input the 3-channel RGB data for testing, and the other is the `4c' version taking as input the 4-channel RGB-T data for testing. Herein we display the results of our approaches with two different relation functions, including dot product and Gaussian function. The one with the Gaussian function performs marginally better than the one using the dot product. We attribute it to the Gaussian function performing relation modeling in a high-dimension space.
Our methods perform comparably against state-of-the-art methods including MMDRNet, GCGLNet, and FEANet. Besides, compared with the recently proposed EGFNet method, CSRPNet (Gaussian) improves the mIoU score by 1.2\%. We also find that several methods tested with the four-channel RGB-T data do not perform better than the version tested with the three-channel RGB data in some cases. We speculate the reason is that these methods ignore the modality gap between the RGB and thermal images, which will adversely affect the segmentation performance.

Furthermore, we compare CSRPNet with the other state-of-the-art methods on the MFNet dataset for daytime and nighttime images. Table~\ref{Table: day_night} reports the corresponding experimental results. CSRPNet (Dot product) and CSRPNet (Gaussian) achieve promising performance for both the daytime and nighttime testing images. Many methods including FuseNet, MFNet, RTFNet-50, and RTFNet-152 ignore the modality gap and thus only obtain limited performance.

\begin{figure}[t!] 
\centering 
\includegraphics[width=1.0\linewidth]{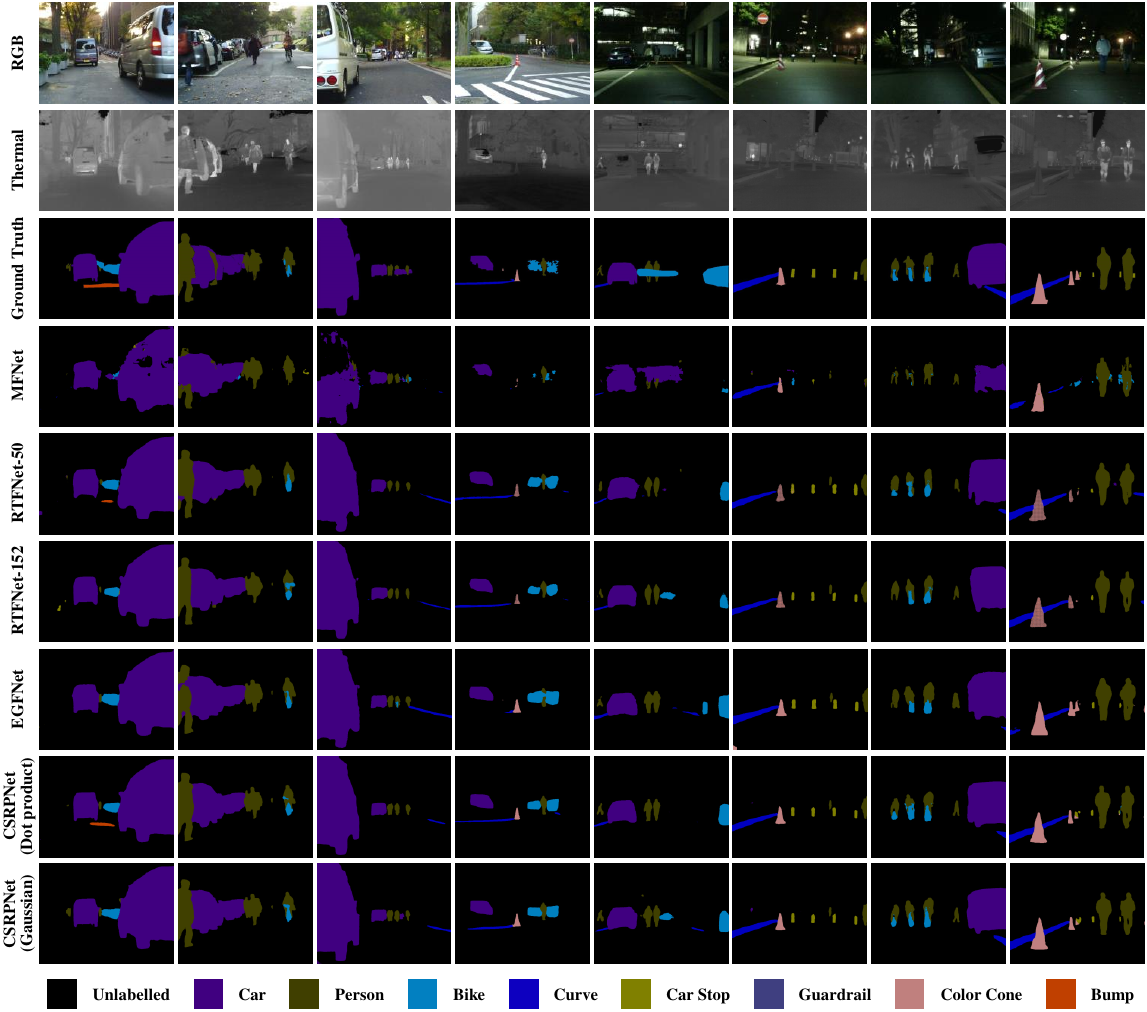} 
\caption{\textbf{Qualitative comparisons between different methods.} Compared with other algorithms, our proposed CSRPNet can predict more precise segmentation results in different lighting conditions.} 
\label{Fig: qualitative_comparison} 
\end{figure}

\vspace{1mm}
\noindent\textbf{The PST900 dataset.}
Seven state-of-the-art methods are involved in the comparison, including Efficient FCN~\cite{EffFCN}, CCNet~\cite{ccnet}, ACNet~\cite{ACNet}, MFNet~\cite{MFNet}, RTFNet~\cite{RTFNet}, and PSTNet~\cite{PSTNet}. The proposed CSRPNet (Dot product) and CSRPNet (Gaussian) obtain mAcc scores of 81.4 and 73.9 and mIoU scores of 73.9 and 75.0, respectively. Our methods perform favorably against the other state-of-the-art algorithms, which demonstrates the effectiveness of our method.

\begin{figure}[t!] 
\centering 
\includegraphics[width=0.9\columnwidth]{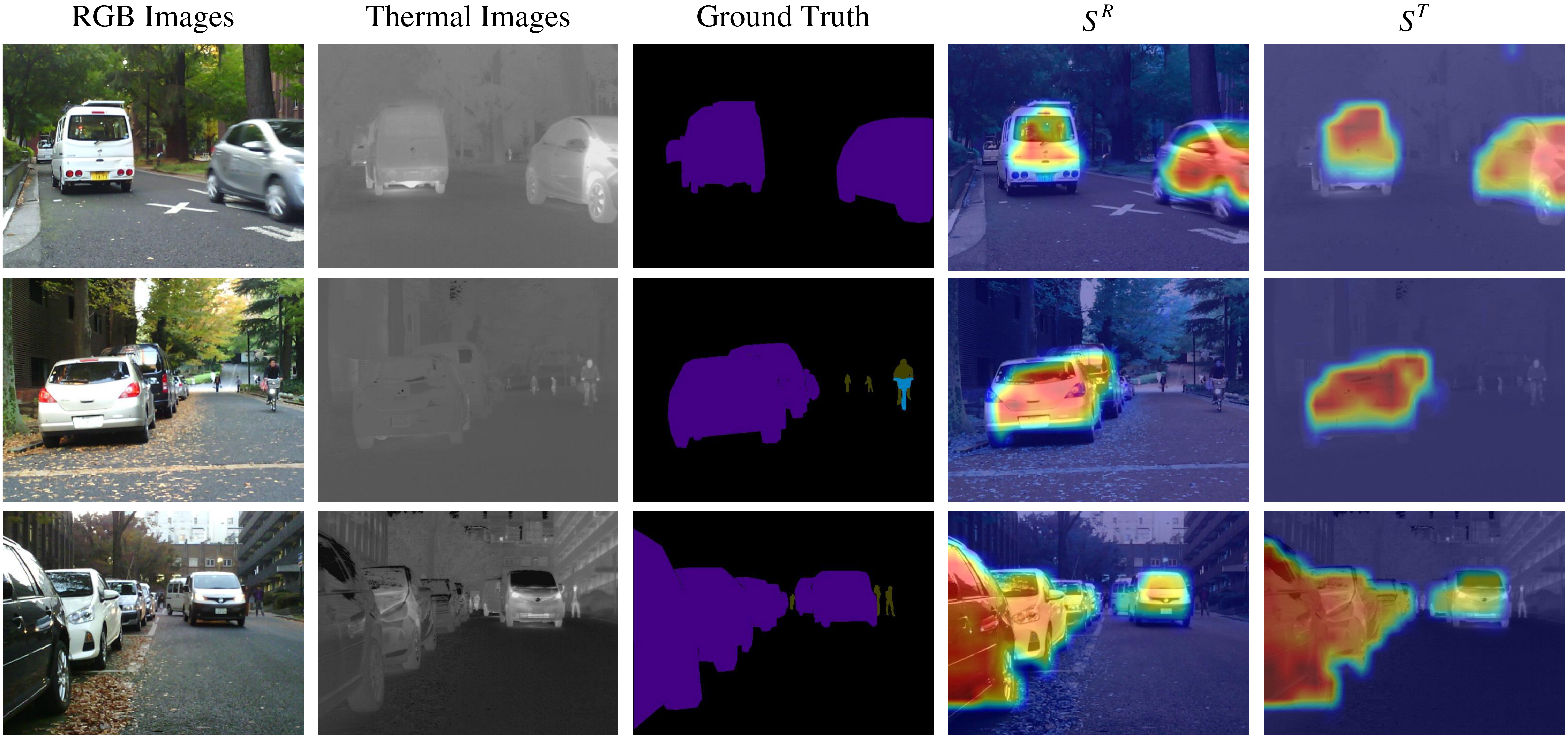} 
\caption{\textbf{Visualization of the modality-shared features calculated via the CSRP module (using dot product) in \emph{layer4}.} The visualized images for the two feature maps are generated by averaging all feature channels. The modality-shared feature of the RGB image $S^{R}$ and the modality-shared feature of the thermal image $S^{T}$ share similar activation distributions and both focus on the foreground objects, which demonstrates that our proposed CSRP module succeeds in capturing the modality-shared information.}
\label{Fig: qualitative_study} 
\end{figure}
\subsection{Qualitative Study}
To obtain more insights into our CSRPNet, we qualitatively compare it with state-of-the-art methods. Figure~\ref{Fig: qualitative_comparison} shows the segmentation results of different methods. The left four image pairs are captured in the daytime, while the right four image pairs are captured in the nighttime. In the daytime scene, the objects with low temperatures are poorly visible in the thermal images, while their texture and color are clearly visible in the RGB images, such as the bump in the first column and the bike in the fourth column. CSRPNet (Dot product) successes recognizing the bump, while the other methods almost cannot. Besides, CSRPNet (Dot product) and CSRPNet (Gaussian) segment the bike more precisely than the MFNet and EGFNet methods. In the night scene, most objects are poorly visible in the RGB images due to insufficient light, while they are clearly visible in the thermal images. Compared with the other methods, CSRPNet (Dot product) and CSRPNet (Gaussian) predict more precise segmentation masks. For example, in the fifth column, the MFNet method misrecognizes several pixels belonging to the person as the car while CSRPNet (Dot product) and CSRPNet (Gaussian) segment the persons precisely. In the seventh column, the EGFNet method misses the bike on the left while CSRPNet (Dot product) and CSRPNet (Gaussian) recognize all the bikes. Besides, the MFNet, RTFNet-50, and RTFNet-152 methods do not use boundaries as auxiliary supervisions. As a result, they cannot process the details of the object boundary well, especially for small objects.

We also visualize the modality-shared features calculated via the CSRP module in \emph{layer4}. Herein dot product is used as the relation function for visualization. Specifically, we visualize two modality-shared feature maps $S^{R}=S^{R}_{ch}+S^{R}_{sp}$ and $S^{T}=S^{T}_{ch}+S^{T}_{sp}$; they contain the modality-shared information captured from the features of the thermal image and the RGB image, respectively. Figure~\ref{Fig: qualitative_study} illustrates the raw RGB-T image pair, ground truth mask, the modality-shared feature of the RGB image $S^{R}$, and the modality-shared feature of the thermal image $S^{T}$. We can observe that $S^{R}$ and $S^{T}$ share similar activation distributions and both focus on the foreground objects. Such an observation validates that the proposed CSRP module effectively captures the modality-shared information, allowing interactive feature fusion across modalities without corrupting the modality-specific information.

%% file: conclusion.tex
In this paper, we propose a channel and spatial relation-propagation network for RGB-T semantic segmentation. Specifically, we propose a channel and spatial relation-propagation (CSRP) module to overcome the modality gap for multi-modality feature fusion. The CSRP module first captures modality-shared features, and then performs feature aggregation based on the captured features, alleviating the issue of contaminating modality-specific information. By plugging the proposed CSRP module into different layers of the two-stream backbone network, we achieve interactive feature fusion without the adverse effect of the modality gap, enabling our model to better leverage the complementary nature between RGB and thermal images. In addition, we propose a dual-path cascaded feature refinement module to make full use of the multi-layer fusion features. The proposed method achieves favorable performance against state-of-the-art methods on the MFNet and PST900 datasets, demonstrating its effectiveness.

%% file: acknowledgements.tex
This research was supported in part by the National Natural Science Foundation of China (No. 62172126), in part by the Shenzhen Research Council (No. JCYJ20210324120202006 and No. JCYJ20210324132212030), and in part by the Major Key Project of PCL (No. PCL2021A03-1 and No. PCL2021A07).